\documentclass[11pt]{article}

\usepackage[utf8]{inputenc}
\usepackage[T1]{fontenc}
\usepackage[margin=1in]{geometry}
\usepackage{amsmath,amssymb}
\usepackage{graphicx}
\usepackage{booktabs}
\usepackage{caption}
\usepackage{microtype}
\usepackage{authblk}
\usepackage{parskip}
\usepackage[numbers]{natbib}
\usepackage[hidelinks]{hyperref}
\usepackage{url}

\graphicspath{{figures/}{../figures/}}

\title{Per-Token Fixed-Point Convergence in Depth-Recurrent Transformers}
\author{Dr.\ Joe Logan}
\affil{Independent researcher \\ \texttt{hello@jlgn.io}}
\date{}

\begin{document}
\maketitle

\begin{abstract}
A depth-recurrent transformer applies a weight-tied core a variable number of
times, and prior work has shown that training with a randomized recursion count
yields one checkpoint usable across a range of inference depths. We ask what
such a model actually computes per token, and measure it directly. On a
135M-class model trained on FineWeb-Edu, the recurrent state converges to a
per-token fixed point: mean successive-output KL divergence falls from
$3.9\times10^{-1}$ at the second loop to $8.5\times10^{-6}$ by the sixteenth,
and per-token state change decays in step. Crucially, this convergence is not
uniform across tokens. The median token converges by loop six, while
approximately 10 percent of tokens continue to update at the training-mean
depth of eight, and mean convergence depth is ordered by token type (whitespace
shallowest, content words deepest). This per-token variation is the central
object of the paper. We show it is directly readable and that reading it
outperforms learning to predict it: a training-free rule that halts each token
once its output stabilizes attains uniform depth-8 quality at 4.94 average loops
(a 38 percent reduction in average depth) and matches uniform depth across the
average-depth range, whereas a linear router trained on convergence labels
harvested from the same model requires nearly full depth and yields no
reduction. The elasticity that makes this possible reproduces here as
background (validation loss decreases monotonically from 3.80 at one loop to
3.20 at eight and remains stable to 32 loops). We report average depth as a FLOP
proxy with a three-point wall-clock bracket rather than a realized speedup, make
no FLOP-matched parity claim, and note that the allocation results are
established at a single scale and seed. The complete study runs on a single
RTX 4090 in approximately 100 GPU-hours.
\end{abstract}

\section{Introduction}

A fixed-depth transformer binds its quality to a single compute operating
point. Serving several quality and cost tiers requires training, storing, and
operating several models, and adapting compute to input difficulty requires
switching among them at inference, which forfeits the prompt cache. Both costs
share a cause: depth, the primary determinant of per-token compute, is fixed
in the weights.

Depth-recurrent transformers, which apply a weight-tied block a variable
number of times \citep{dehghani2019universal,geiping2025recurrent}, relax this
constraint. A single parameter set can in principle operate at many depths, so
one training run can cover a range of compute budgets, and the depth can be
selected per request or per token at inference. Whether this holds in practice
depends on two properties that prior work often conflates. The first is depth
robustness: whether the model produces useful output at depths for which it
was not specifically tuned. The second is depth allocation: given a robust
backbone, how much depth each token should receive, and whether the allocation
policy must be learned jointly with the model.

Prior work leaves a gap at this seam. Per-token early exit on non-shared
stacks stalled on the key-value cache and on batching: a token that
permanently skips deep layers writes no keys or values there, so later tokens
have nothing to attend to, and the published methods sacrifice one of input
adaptivity, cache correctness, or compute savings
\citep{schuster2022calm,delcorro2023skipdecode,elhoushi2024layerskip}.
Mixture-of-Recursions \citep{bae2025mor} trains the depth router and the
compute budget together at pretraining time, which fixes the budget in the
weights; the router saturates and cannot be retargeted afterward. Huginn
\citep{geiping2025recurrent} shows that randomized-depth pretraining produces a
depth-robust backbone at 3.5B parameters, but it sets a single global depth by
hand at inference and does not characterize its own emergent stop rule or test
whether a learned policy could beat it. LoopFormer \citep{jeddi2026loopformer},
concurrent with this work, improves the training objective for elastic depth
but allocates a single global budget per sequence and names per-token budgeting
as a limitation. Jointly trained halting has an established history:
adaptive-computation methods that learn a halting policy alongside the model
\citep{graves2016act,banino2021pondernet} are known to collapse to degenerate
depths and to require delicate tuning of a compute prior
\citep{csordas2026utmemory}. In each case, depth allocation is either fixed
manually, applied globally per sequence, or learned jointly with the model, and
in none of these lines is the per-token computation of a trained elastic model
measured directly. We take that measurement as the starting point: we hold a
randomized-depth backbone fixed, characterize how its recurrent state evolves
per token, and ask what the simplest policy that reads this behavior achieves
relative to a policy trained to predict it.

We separate the two problems and address them with distinct methods. Depth
robustness is a pretraining property, obtained here with a minimal recipe: the
recursion count is sampled randomly at each optimizer step and the model is
trained with standard cross-entropy, without a halting unit, ponder loss, or
FLOP penalty. Depth allocation is then a post-hoc policy on the frozen
backbone. Such a policy is inexpensive to construct, can be replaced or
retuned to any budget without modifying the weights, and admits fair
comparison against alternatives because the backbone is held fixed.

Our contributions, in order of strength:

\begin{enumerate}
\item A direct per-token measurement of the recurrence. The state converges to
  a per-token fixed point (mean successive-output KL falls from
  $3.9\times10^{-1}$ to $8.5\times10^{-6}$ between loops 2 and 16, with state
  change decaying in step), and the convergence depth varies across tokens: the
  median token converges by loop six while approximately 10 percent are still
  updating at the training-mean depth, ordered by token type. To our knowledge
  this per-token convergence structure has not previously been characterized in
  a depth-recurrent language model.
\item A demonstration that this variation is directly exploitable and that
  reading it outperforms learning it. A training-free convergence exit attains
  uniform depth-8 quality at 4.94 average loops (38 percent lower average depth)
  and matches uniform depth across the average-depth range, whereas a linear
  router trained on convergence labels harvested from the same model requires
  nearly full depth and yields no reduction. The training-free rule
  outperforming the learned one, reproduced at two scales, is the clearest
  allocation result.
\item A hypothesis, supported by two internal data points and one external
  consistency check \citep{geiping2025recurrent}, that saturation depth tracks
  the training-time loop distribution: our mean-4 model saturates at depth 4 and
  our mean-8 model at depth 8.
\end{enumerate}

We reproduce depth elasticity as the setting for these results, not as a
contribution. One randomized-depth checkpoint serves depths 1~to~8 with
monotone quality and stable extrapolation to four times the training-mean
depth. This property was established by prior work at larger scale
\citep{geiping2025recurrent,jeddi2026loopformer}; here it is the fixed backbone
on which the measurement and allocation results are obtained.

We state the scope limits explicitly. We report no FLOP-matched parity against
a standard transformer; the comparison to SmolLM2-135M \citep{allal2025smollm2}
is a ceiling reference at a 150-fold token disadvantage, not a matched baseline.
We report no realized wall-clock speedup, only a bracket between the FLOP
proxy and the best case achievable under depth-wise batching. Quality does not
improve beyond the training-mean depth, contrary to our initial hypothesis.

\section{Related work}

\paragraph{Early exit and its failure modes.}
CALM \citep{schuster2022calm} exits tokens early from a deep non-shared stack
and copies the exit-layer state upward to approximate the missing keys and
values; SkipDecode \citep{delcorro2023skipdecode} makes exit depth a fixed
function of position to keep the cache correct, at the price of input
adaptivity; LayerSkip \citep{elhoushi2024layerskip} recovers exactness by
verifying drafts with the full model, at the price of eventually running every
layer anyway. The recurring pattern is a trilemma between input-adaptivity,
key-value correctness, and actual compute savings. Reported speedups are also
often batch-1 latency numbers that shrink under continuous batching; TIDE
\citep{whitfield2026tide} reports 98~to~99 percent of tokens exiting early for
single-digit percent wall-clock gain, and a recent serving system for batched
early exit recovers 2~to~12 percent throughput \citep{rao2025drex}. This gap
between FLOP accounting and wall-clock reality bears directly on how we scope
the efficiency result in Section~\ref{sec:allocation}. Weight-tied recursion
sidesteps the trilemma structurally: every depth uses the same weights, so
tokens at different depths occupy the same batch, and the cache story is
per-recursion rather than per-layer.

\paragraph{Budgeted routing.}
Mixture-of-Depths \citep{raposo2024mod} and Mixture-of-Recursions
\citep{bae2025mor} route tokens to variable compute with expert-choice top-$k$
selection, which fixes the batching and budget problems that constrained early
exit: the compute per block is exact and the graph is static. The cost is that
the router is co-trained and the budget is set at pretraining. Bae et al. state
the limitation directly: the router saturates, capacity cannot be re-targeted
after training, and some tokens are never selected. Expert-choice selection is
also non-causal at decode time and needs an auxiliary predictor to patch. Our
post-hoc policy targets exactly this limitation: because the backbone is trained
with no router at all, any budget can be dialed in afterward, and the allocation
rule is replaceable.

\paragraph{Depth recurrence and elastic-depth training.}
Universal Transformers \citep{dehghani2019universal} established the weight-tied
recurrent core with ACT-style halting; Huginn \citep{geiping2025recurrent}
scaled the design to 3.5B parameters by discarding learned halting entirely,
sampling the recursion count per training step, and reading depth as a test-time
knob; Ouro \citep{zhu2025ouro} pretrained looped models at trillion-token scale
with an entropy-regularized depth objective. Geiping et al. also observe an
emergent convergence-based stop rule (exit when successive iterates stop
changing) but do not characterize it against alternatives. Concurrent with this
work, LoopFormer \citep{jeddi2026loopformer} trains a 1B-parameter looped model
with a shortcut-consistency objective over families of step schedules, so that
one checkpoint serves a range of user-specified compute budgets, and reports
gains in perplexity and zero-shot reasoning as the budget grows. LoopFormer
allocates compute with a single global budget per sequence and lists per-token,
instance-adaptive budgeting as an open limitation. Our work addresses exactly
that limitation, on a frozen backbone and without a new training objective: we
hold the elastic model fixed and study which per-token allocation policy to run
on top of it. The two results are complementary. A better elastic backbone and
a better post-hoc allocation rule compose, and the allocation policies we
evaluate apply to any depth-robust looped model, including theirs.

The halting literature trains the halting policy jointly with the model and
inherits collapse and tuning pathologies: ACT \citep{graves2016act} requires a
per-task search over its ponder cost, PonderNet \citep{banino2021pondernet} is
sensitive to its compute prior, and a documented failure mode is that most runs
fall into a shallow-halt equilibrium that training never escapes
\citep{csordas2026utmemory}. This history motivates our central design decision:
the pretraining objective never sees a depth policy, and every allocation rule
we evaluate is applied to the same frozen backbone afterward.

\section{Method}

\subsection{Architecture}

The model has three parts: a prelude of unique layers that reads the input into
latent space, a weight-tied core applied a variable number of times, and a coda
of unique layers feeding the language-model head. The layers are Llama-style
\citep{touvron2023llama}: RMSNorm \citep{zhang2019rmsnorm}, SwiGLU
\citep{shazeer2020glu}, rotary position embeddings \citep{su2024roformer},
grouped-query attention \citep{ainslie2023gqa}, tied input and output
embeddings, no biases. The recurrence follows \citet{geiping2025recurrent}, and
each design choice below exists to prevent a specific documented failure.

The prelude output $e$ is computed once per sequence. The recurrent state is
initialized from noise, $s_0 \sim \mathcal{N}(0, \sqrt{2/5})$, not from $e$;
together with input re-injection this pushes the core toward computing an
input-conditioned fixed point rather than memorizing an iteration count. Each
loop re-injects the input by concatenation,
$s_{i+1} = \mathrm{core}(\mathrm{adapter}([e, s_i]))$, where the adapter is a
single linear map from $2d$ to $d$; without re-injection the core can drift
away from the input over many iterations. The core blocks use sandwich
normalization (a post-norm inside each residual branch in addition to the
pre-norm), which Geiping et al. report as a fix for hidden-state collapse in
looped training. Configuration for both scales is in Table~\ref{tab:config}.

\subsection{Training}

Per optimization step we sample one recursion count,
$r \sim \mathrm{round}(\mathrm{LogNormal}(\log \bar{r}, 0.5))$ clamped to
$[1, r_{\max}]$, and apply it to the whole step. The loss is plain
cross-entropy at the final unrolled state. There is no halting unit, no ponder
loss, and no FLOP penalty; the halting pathologies reviewed in Section 2 are
the reason. Randomizing the depth is the entire mechanism by which the model
becomes usable at every depth: each depth is visited during training in
proportion to the sampling distribution, so no single unroll length is
privileged and none is foreign.

We backpropagate through only the last $k$ iterations (truncated backprop) to
bound activation memory; earlier iterations run under no-grad. S0 uses
$\bar{r}=4$, $r_{\max}=16$, $k=4$; S1 uses $\bar{r}=8$, $r_{\max}=32$, $k=8$.
Optimization is AdamW \citep{loshchilov2019adamw} with weight decay on matrix
parameters only, gradient clipping at 1.0, and a cosine schedule with warmup,
run to completion; every headline number in this paper comes from the fully
annealed final checkpoint, not from a mid-schedule one.

\subsection{Post-hoc allocation policies}

All three policies run on the frozen backbone. None of them updates a weight of
the language model.

\begin{itemize}
\item \textbf{A0, uniform depth.} Apply the same loop count $r$ to every token;
  sweep $r$. This is the baseline every adaptive policy must beat, and it is
  also the per-request operating point: a single depth for the entire request.
\item \textbf{A1, convergence exit (training-free).} After each loop $i$,
  compute $\mathrm{KL}(p_i \,\|\, p_{i-1})$ between successive output
  distributions per token; once it falls below a threshold $\epsilon$, freeze
  that token and carry its output forward. Sweeping $\epsilon$ traces an
  (average-depth, loss) curve. The policy has no parameters and requires one
  forward hook.
\item \textbf{A2, oracle-labeled router.} Harvest labels by running tokens to a
  high loop count and recording, per token, whether the argmax prediction still
  changes after loop $i$. Train a linear probe on the state $s_i$ to predict
  this, then use it as the stop rule. The labels come from the frozen model
  itself, so no search or external supervision is needed. A2 exists to test
  whether learning adds anything over the free heuristic; at both our scales the
  answer is no.
\end{itemize}

\section{Experimental setup}

\paragraph{Scales.}
We study two scales. S0 has 12.9M unique non-embedding parameters and trains on
400M tokens of TinyStories \citep{eldan2023tinystories}; it serves as a
controlled, fast-iteration setting for architecture validation and replication,
and we do not draw quantitative claims from it. S1 has 29.0M unique
non-embedding parameters (57.3M total) and trains on 12.9B tokens of FineWeb-Edu
\citep{penedo2024fineweb} with a single seed. Its shape follows SmolLM2-135M
\citep{allal2025smollm2}, with the middle layers replaced by the weight-tied
core, so its per-token compute at the training-mean depth is comparable to a
135M dense model. S0 has two of three planned seeds complete, so we report the
observed two-seed spread and make no seed-variance claim. Both scales use the
SmolLM2 tokenizer (49{,}152 entries), giving an identical data pipeline and
evaluation harness across scales.

\paragraph{Eval conventions.}
Two evaluation harnesses appear in the results. Loss-versus-loops sweeps use
noise state initialization (the training path) and 64 evaluation windows of
length 2048. The allocation comparison uses zero state initialization, for
reproducible per-token exit decisions, and 128 windows. The two conventions
give the same model an $r=8$ loss of 3.20 (noise, sweep) and 3.19 (zero,
allocation A0); the difference is 0.013 nats and comes from the initialization
and window count. We quote 3.20 as the headline elasticity number and use 3.19
only as the internal reference for the A0/A1/A2 comparison, which is
self-consistent because all three policies are measured in the same harness.
The SmolLM2 reference is measured in the sweep harness, matching the 3.20
convention.

\paragraph{Hardware and cost.}
S1 training ran on one RTX 4090 for approximately 100 GPU-hours (12.9B tokens at
24k tokens per second, 262k-token effective steps). All evaluation, including
the post-hoc allocation study, runs on a single GPU. Appendix~\ref{app:stability}
documents a mixed-precision attention bug we encountered and fixed;
practitioners running mixed-precision rotary attention are likely to hit the
same one.

\section{Results}

\subsection{Elasticity (the setting)}

We first confirm that the backbone is depth-elastic, since the measurement and
allocation results in the following subsections require it. This property is not
a contribution of this work; it establishes the fixed model on which the rest is
measured. Table~\ref{tab:elasticity} gives S1 validation loss as a function of
inference loop count for the final checkpoint. Loss falls monotonically from
3.80 at one loop to 3.20 at eight, then stays flat: $r=8$, 16, and 32 are 3.202,
3.203, and 3.203. The useful range spans a 16-fold difference in per-token
compute ($r=1$ versus $r=16$) with a 0.60-nat quality range across it, all from
one set of weights chosen at inference time. Running four times deeper than the
training mean does not degrade quality. The same shape held at every checkpoint
we swept during training (Appendix~\ref{app:elasticity}), so it is a stable
property of the model, not of a particular point in the schedule.

The fixed-depth control quantifies the contrast. A model trained at a constant
four loops (S0) is sharply peaked at its training depth: loss 1.97 at $r=4$ but
6.01 at $r=1$ and 2.50 at $r=16$. The randomized-depth model at the same scale
is flat and low across $r=4, 8, 16$ (1.886, 1.883, 1.883), and it is slightly
better than the control even at the control's own specialized depth. Randomized
depth confers robustness that fixed depth does not, at no cost at the depth for
which the fixed model was tuned.

\begin{figure}[t]
\centering
\includegraphics[width=\linewidth]{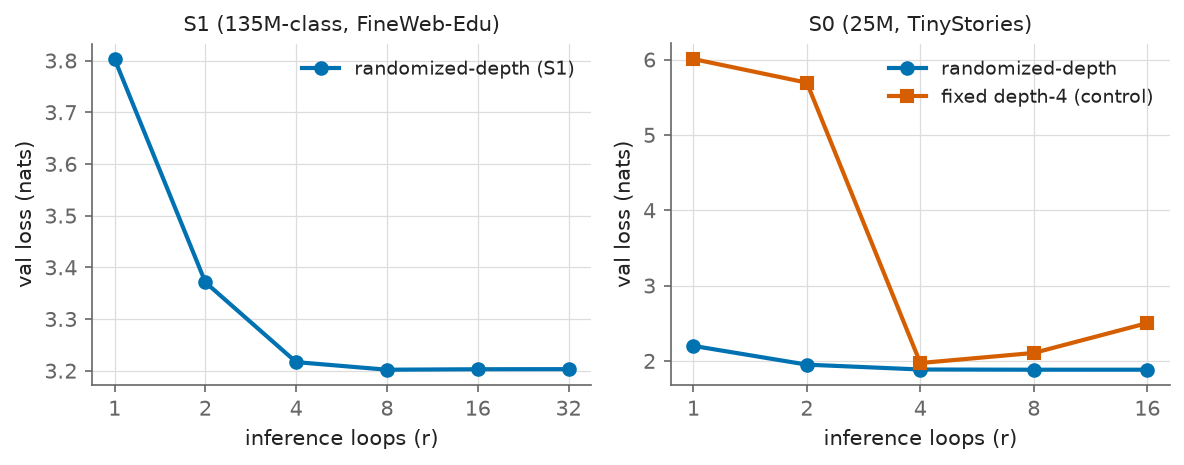}
\caption{Validation loss as a function of inference loop count. Left: the S1
model (135M-class compute, FineWeb-Edu) falls monotonically from 3.80 at one
loop to 3.20 at eight and stays flat to 32 loops. Right, S0 (31.8M total
parameters, TinyStories): the randomized-depth model is flat and low across
depths, while a model trained at a fixed depth of four is sharply peaked at its
training depth and degrades at every other depth. Randomized depth confers
robustness that fixed depth does not.}
\label{fig:elasticity}
\end{figure}

\subsection{Why saturation at the training mean}
\label{sec:saturation}

The flat region beyond $r=8$ admits two readings: the loop could be spinning
without effect, or the computation could be genuinely finished. These are
distinguishable by measurement, and the distinction matters because the two
readings imply different things about the architecture.
Table~\ref{tab:convergence} reports, per loop, the mean successive-output KL,
the mean state change $\|s_i - s_{i-1}\|$, and the fraction of tokens still
moving (KL above $10^{-3}$). Both KL and state change decay smoothly toward zero
together: KL is $3.9\times10^{-1}$ at loop 2, $5.4\times10^{-4}$ at loop 8, and
$8.5\times10^{-6}$ at loop 16. The state reaches a fixed point. Loss saturates
at the training mean because the computation is finished there, not because the
loop is broken. This is the behavior the architecture was designed for (noise
initialization plus input re-injection push the core toward a contraction), and
it explains why extrapolated depths are safe: iterating past convergence leaves
a converged state where it is.

The decay is not uniform across tokens. At loop 7, 33 percent of tokens are
still moving; at loop 8, 10 percent; by loop 16, 0.1 percent. The median token
has converged by about loop 6, but a tail of roughly one token in ten needs the
full training-mean depth. Uniform depth incurs the tail's cost on every token.
That tail is the allocation opportunity, and Section~\ref{sec:allocation} is
downstream of this measurement: we measured the tail first and then tested
whether a policy could exploit it.

\begin{figure}[t]
\centering
\includegraphics[width=\linewidth]{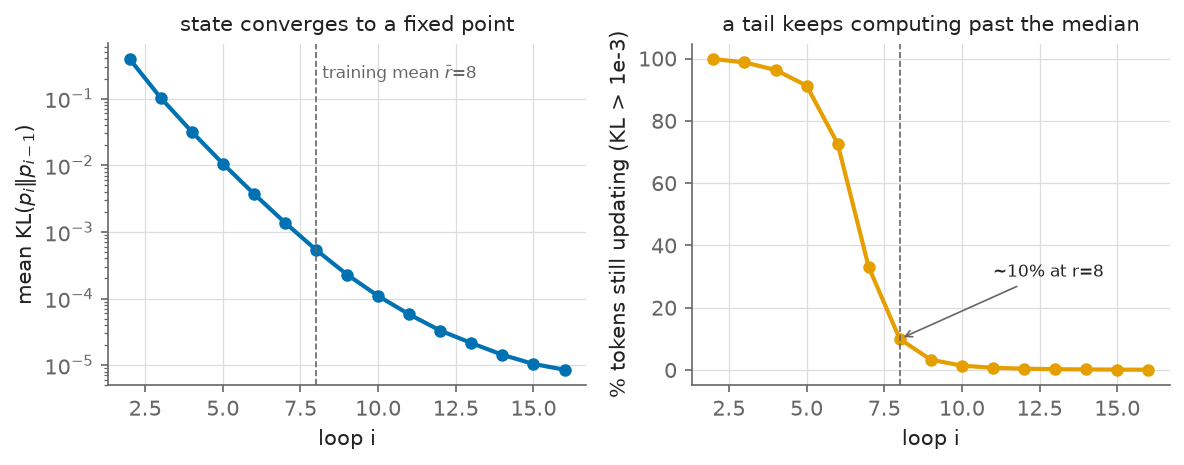}
\caption{Per-loop convergence at S1. Left: mean successive-output KL decays
log-linearly to $8.5\times10^{-6}$ by loop 16; the state reaches a fixed point.
Right: the fraction of tokens still updating (KL above $10^{-3}$) drops from near
100 percent to about 10 percent at the training-mean depth of eight, with the
median token converged by loop six. The dashed line marks the training-mean
depth. The tail of late-converging tokens is what per-token allocation
exploits.}
\label{fig:convergence}
\end{figure}

\subsection{Allocation}
\label{sec:allocation}

Table~\ref{tab:allocation} gives loss at matched average depth for the three
policies on the final model (128 windows). The uniform depth-8 quality bar is
3.189. The convergence exit A1 reaches that quality at 4.94 average loops, a 38
percent reduction in average depth, and it matches or exceeds uniform depth
across the range. At average depth 3 the margin is clear (3.231 against 3.272);
at average depths 4~through~7 the two are within 0.002~to~0.003 nats, with A1
never higher. At these evaluation sizes we claim that A1 matches uniform depth
at matched average depth and outperforms it at the shallow end, without claiming
strict dominance at every point. The mechanism is the one measured in
Section~\ref{sec:saturation}: A1 halts the converged majority early and
allocates the reclaimed loops to tokens whose state is still evolving.

The trained router A2 needs average depth 7.99 to reach the same quality,
yielding no saving, and underperforms A1 across the range. This reproduces and
strengthens the S0 result, in which A1 matched uniform depth and A2
underperformed. The explanation is direct: the convergence exit observes the
model's per-token state dynamics, whereas the probe predicts them from a single
state snapshot and is a less accurate estimator of the same quantity. At S0 the
two policies had little to separate them because the toy model's useful-depth
range was narrow; at S1 the wider range exposes the difference. A policy that
must be trained, calibrated, and re-tuned per budget loses to a policy with zero
parameters.

Average depth is a FLOP proxy, and we scope the efficiency result accordingly.
In our teacher-forced masking harness, A1 and A2 loop every token to $r_{\max}$
and mask the frozen ones, so they skip no work and realized throughput does not
beat uniform depth. A matched-depth wall-clock timing (uniform $r=8$ against
uniform $r=5$, the approximate A1 average) gives a 1.48~to~1.51 times speedup
across batch sizes 1~to~8, which is the best case achievable if depth-wise
batching packed exited tokens perfectly. The three-point bracket is therefore:
38 percent by average depth, about 1.5 times best-case wall-clock, and about 0
percent realized in our harness. Unlike layer-skip early exit, where a 3 times
FLOP saving collapsed to single-digit wall-clock because the skips are ragged
per-layer holes in a batch, the recursive saving is in whole forward passes of a
shared-weight core, so exited tokens free schedulable work and the FLOP proxy
and best-case wall-clock stay close. Closing the remaining gap is a systems
problem (depth-wise continuous batching, demonstrated by \citet{bae2025mor})
that we did not build.

\begin{figure}[t]
\centering
\includegraphics[width=\linewidth]{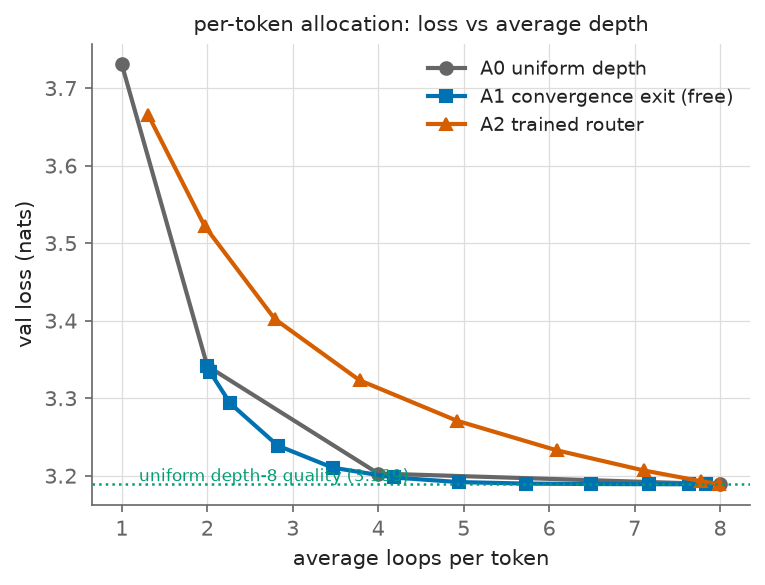}
\caption{Loss against average loops per token for the three policies (S1 final).
The dotted line is uniform depth-8 quality. The training-free convergence exit
A1 reaches that quality at about five average loops and tracks or exceeds
uniform depth A0 across the range. The trained router A2 needs close to eight
average loops to reach the same quality and underperforms A1 across the range.}
\label{fig:allocation}
\end{figure}

\subsection{Token-level behavior}

Depth demand tracks token type. Mean exit depth at S1 orders as space 3.58,
numeral 3.81, punctuation 4.16, word 4.25. At S0 the order was space,
punctuation, word, numeral, with numerals deepest. The ordering of the two
content classes flips between scales. We attribute this to distribution shift:
TinyStories contains simple recurring arithmetic that rewards depth on numerals,
whereas FineWeb-Edu numerals are often rare isolated tokens the model cannot
resolve further, and words carry the compositional load. The ordering was stable
under a four-fold increase in evaluation windows, so it is a dataset effect, not
sampling noise. The general pattern, structural tokens exiting early and content
tokens computing longer, matches what the adaptive-depth literature has reported
since ACT \citep{graves2016act}, here obtained without any learned halting.

\begin{figure}[t]
\centering
\includegraphics[width=\linewidth]{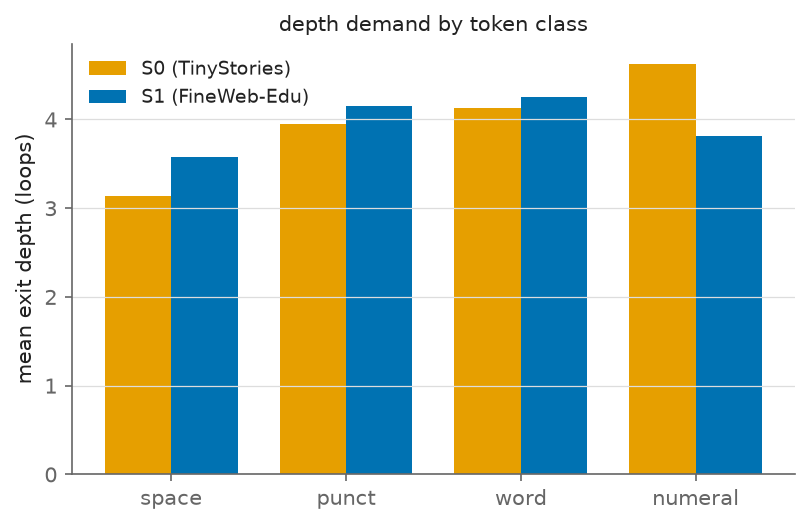}
\caption{Mean exit depth by token class at both scales. Structural tokens
(whitespace) exit shallowest and content tokens compute longest at both scales.
The relative order of numerals and words flips between TinyStories (numerals
deepest) and FineWeb-Edu (words deepest), which we attribute to distribution
shift. The ordering was stable under a four-fold increase in evaluation
windows.}
\label{fig:tokenclass}
\end{figure}

\subsection{Ceiling reference}

Evaluated under our S1 harness, SmolLM2-135M scores 2.66 nats against the
recursive model's 3.20 at $r=8$, a gap of 0.54 nats. This is a ceiling
reference, not a matched baseline, for two reasons. First, SmolLM2-135M trained
on roughly 2T tokens, about 150 times our budget, so a gap is expected from the
token count alone. Second, SmolLM2 trained on FineWeb-Edu while our evaluation
set is FineWeb-Edu validation, so plausible train-eval overlap inflates its
score and widens the measured gap. Both effects widen the gap rather than narrow
it.

\subsection{The per-request cost curve}

For the deployment framing, Table~\ref{tab:cost} restates the elasticity sweep
as a cost curve: quality, measured tokens per second at batch 1 on one RTX 4090,
and relative throughput at each uniform depth. From one checkpoint, an operator
can serve depth 2 at 2.7 times the throughput of depth 8 for a 0.17-nat quality
cost, or any point between, without a model switch or cache invalidation. These
numbers are one GPU class and one implementation; they are included to make the
product shape concrete, not as a serving benchmark.

\section{Discussion and limitations}

\paragraph{Useful depth is determined at training time.}
Our hypothesis that useful depth would extend beyond the training mean at larger
scale did not hold: the model saturates at the training-mean depth and gains
nothing beyond it. Across our two scales the pattern is exact: trained at mean 4,
saturates at 4; trained at mean 8, saturates at 8. Huginn, trained at mean 32,
reports useful depth to 32 and beyond \citep{geiping2025recurrent}, which is
consistent with the same rule. Two internal points and one external consistency
check do not establish a law, and we have not tested a wider training
distribution at fixed scale, so we state this as the paper's main open
hypothesis: a recursive model uses depth up to its training mean and no further.
If it holds, it is a design lever: the saturation depth, and with it the width
of the product's quality-compute range, is chosen by the practitioner through
the training loop distribution. The convergence measurement in
Section~\ref{sec:saturation} shows the model is not failing to use extra depth;
it has finished computing.

\paragraph{Baselines and seeds.}
We make no FLOP-matched parity claim. A matched vanilla transformer costs the
same as the recursive run at iso-FLOP by construction, and there is no
legitimate discounted version: a token-starved short vanilla would lose because
it saw less data, not because it is architecturally worse, so we rejected that
comparison. Parity is the natural next experiment, and the one on which any
deployment argument most depends. S1 is a single seed. The S0 seed study has two
of three seeds complete, with a maximum spread of 0.03 nats across the sweep and
0.011 nats at $r=16$ between the two; we report this as observed and make no
seed-variance claim.

\paragraph{Wall-clock.}
Realizing the average-depth saving as throughput requires depth-wise continuous
batching, as demonstrated by \citet{bae2025mor}. We did not build it, and we
report the three-point bracket rather than an end-to-end speedup. We note that
the structural situation is more favorable than for layer-skip early exit, for
the reason given in Section~\ref{sec:allocation}, but a measured serving result
is future work.

\paragraph{Scale.}
All results are at 135M-class compute. The literature suggests the interesting
depth effects strengthen with scale and task difficulty (Geiping et al. report
reasoning-heavy tasks improving out to depth 64 while easy benchmarks saturate
near 8), so our average-loss results on web text plausibly understate what
per-token allocation offers on harder distributions. We have not tested this;
behavior at 1B parameters and above is unknown.

\paragraph{What we did not solve.}
The key-value cache story for per-token allocation at decode time is inherited
from the recursion literature (per-recursion caching or depth-1 sharing
\citep{bae2025mor}) and was not stress-tested here; our evaluation is
teacher-forced. Downstream task evaluation at each depth, which would show
whether reasoning-flavored capabilities track depth differently from average
loss, was out of budget and remains open.

\section{Conclusion}

The recurrent state of a randomized-depth transformer converges to a per-token
fixed point, and the depth at which it converges varies across tokens: the
median token converges well before the training-mean depth while a minority
continues to update at it. This variation is directly readable from the
successive outputs, and reading it outperforms learning to predict it. A
training-free convergence exit attains uniform depth-8 quality at 4.94 average
loops and matches uniform depth across the average-depth range, with an
advantage at shallow depths, whereas a router trained on labels from the same
model requires close to full depth and provides no reduction. These allocation
results hold at a single 135M-class scale on one seed, with average depth
reported as a FLOP proxy rather than measured throughput and without a
FLOP-matched dense baseline; the underlying convergence measurement is the more
scale-independent finding. The depth elasticity on which all of this rests
reproduces here from prior work rather than being contributed by it. Directions
for future work are establishing the training-mean saturation pattern as a
general property, converting the average-depth reduction into measured
throughput under depth-wise batching, evaluating downstream capabilities as a
function of depth, and testing whether the per-token convergence structure holds
at larger scale. The complete study reproduces on a single RTX 4090 in
approximately 100 GPU-hours.

\section*{Reproducibility statement}

The training and evaluation code, configuration files, the experiment log, and
the final S1 checkpoint are available at
\url{https://github.com/jlognn/depth-recurrent-convergence}. Every number
reported in this paper is recorded in the released log alongside the
configuration and evaluation harness that produced it. The S1 model trains in
approximately 100 GPU-hours on a single 24GB GPU.

\bibliographystyle{plainnat}
\bibliography{references}

@article{ainslie2023gqa,
  title={{GQA}: Training Generalized Multi-Query Transformer Models from Multi-Head Checkpoints},
  author={Ainslie, Joshua and Lee-Thorp, James and de Jong, Michiel and Zemlyanskiy, Yury and Lebr\'on, Federico and Sanghai, Sumit},
  journal={arXiv preprint arXiv:2305.13245},
  year={2023}
}

@article{allal2025smollm2,
  title={{SmolLM2}: When Smol Goes Big -- Data-Centric Training of a Small Language Model},
  author={Allal, Loubna Ben and Lozhkov, Anton and Bakouch, Elie and others},
  journal={arXiv preprint arXiv:2502.02737},
  year={2025}
}

@article{bae2025mor,
  title={Mixture-of-Recursions: Learning Dynamic Recursive Depths for Adaptive Token-Level Computation},
  author={Bae, Sangmin and Fisch, Adam and Harutyunyan, Hrayr and others},
  journal={arXiv preprint arXiv:2507.10524},
  year={2025}
}

@article{banino2021pondernet,
  title={{PonderNet}: Learning to Ponder},
  author={Banino, Andrea and Balaguer, Jan and Blundell, Charles},
  journal={arXiv preprint arXiv:2107.05407},
  year={2021}
}

@article{csordas2026utmemory,
  title={Universal Transformers Need Memory: Depth-State Trade-offs in Recurrent Language Models},
  author={Csord\'as, R\'obert and others},
  journal={arXiv preprint arXiv:2604.21999},
  year={2026}
}

@inproceedings{dehghani2019universal,
  title={Universal Transformers},
  author={Dehghani, Mostafa and Gouws, Stephan and Vinyals, Oriol and Uszkoreit, Jakob and Kaiser, Lukasz},
  booktitle={International Conference on Learning Representations (ICLR)},
  year={2019},
  note={arXiv:1807.03819}
}

@article{delcorro2023skipdecode,
  title={{SkipDecode}: Autoregressive Skip Decoding with Batching and Caching for Efficient LLM Inference},
  author={Del Corro, Luciano and Del Giorno, Allie and Agarwal, Sahaj and Yu, Bin and Awadallah, Ahmed and Mukherjee, Subhabrata},
  journal={arXiv preprint arXiv:2307.02628},
  year={2023}
}

@article{eldan2023tinystories,
  title={{TinyStories}: How Small Can Language Models Be and Still Speak Coherent English?},
  author={Eldan, Ronen and Li, Yuanzhi},
  journal={arXiv preprint arXiv:2305.07759},
  year={2023}
}

@inproceedings{elhoushi2024layerskip,
  title={{LayerSkip}: Enabling Early Exit Inference and Self-Speculative Decoding},
  author={Elhoushi, Mostafa and Shrivastava, Akshat and Liskovich, Diana and others},
  booktitle={Annual Meeting of the Association for Computational Linguistics (ACL)},
  year={2024},
  note={arXiv:2404.16710}
}

@article{geiping2025recurrent,
  title={Scaling up Test-Time Compute with Latent Reasoning: A Recurrent Depth Approach},
  author={Geiping, Jonas and McLeish, Sean and Jain, Neel and others},
  journal={arXiv preprint arXiv:2502.05171},
  year={2025}
}

@article{graves2016act,
  title={Adaptive Computation Time for Recurrent Neural Networks},
  author={Graves, Alex},
  journal={arXiv preprint arXiv:1603.08983},
  year={2016}
}

@article{jeddi2026loopformer,
  title={{LoopFormer}: Elastic-Depth Looped Transformers with Shortcut Consistency},
  author={Jeddi, Ahmad and others},
  journal={arXiv preprint arXiv:2602.11451},
  year={2026}
}

@inproceedings{loshchilov2019adamw,
  title={Decoupled Weight Decay Regularization},
  author={Loshchilov, Ilya and Hutter, Frank},
  booktitle={International Conference on Learning Representations (ICLR)},
  year={2019},
  note={arXiv:1711.05101}
}

@article{penedo2024fineweb,
  title={The {FineWeb} Datasets: Decanting the Web for the Finest Text Data at Scale},
  author={Penedo, Guilherme and Kydl\'i\v{c}ek, Hynek and Lozhkov, Anton and others},
  journal={arXiv preprint arXiv:2406.17557},
  year={2024}
}

@article{rao2025drex,
  title={{DREX}: Dynamic Rebatching for Early-Exit Inference Serving},
  author={Rao, D. and others},
  journal={arXiv preprint arXiv:2512.15705},
  year={2025}
}

@article{raposo2024mod,
  title={Mixture-of-Depths: Dynamically Allocating Compute in Transformer-Based Language Models},
  author={Raposo, David and Ritter, Sam and Richards, Blake and Lillicrap, Timothy and Humphreys, Peter Conway and Santoro, Adam},
  journal={arXiv preprint arXiv:2404.02258},
  year={2024}
}

@inproceedings{schuster2022calm,
  title={Confident Adaptive Language Modeling},
  author={Schuster, Tal and Fisch, Adam and Gupta, Jai and others},
  booktitle={Advances in Neural Information Processing Systems (NeurIPS)},
  year={2022},
  note={arXiv:2207.07061}
}

@article{shazeer2020glu,
  title={{GLU} Variants Improve Transformer},
  author={Shazeer, Noam},
  journal={arXiv preprint arXiv:2002.05202},
  year={2020}
}

@article{su2024roformer,
  title={{RoFormer}: Enhanced Transformer with Rotary Position Embedding},
  author={Su, Jianlin and Ahmed, Murtadha and Lu, Yu and Pan, Shengfeng and Bo, Wen and Liu, Yunfeng},
  journal={Neurocomputing},
  year={2024},
  note={arXiv:2104.09864}
}

@article{touvron2023llama,
  title={{LLaMA}: Open and Efficient Foundation Language Models},
  author={Touvron, Hugo and Lavril, Thibaut and Izacard, Gautier and others},
  journal={arXiv preprint arXiv:2302.13971},
  year={2023}
}

@article{whitfield2026tide,
  title={{TIDE}: Token-Informed Depth Execution for Efficient Inference},
  author={Whitfield, S. and others},
  journal={arXiv preprint arXiv:2603.21365},
  year={2026}
}

@inproceedings{zhang2019rmsnorm,
  title={Root Mean Square Layer Normalization},
  author={Zhang, Biao and Sennrich, Rico},
  booktitle={Advances in Neural Information Processing Systems (NeurIPS)},
  year={2019},
  note={arXiv:1910.07467}
}

@article{zhu2025ouro,
  title={Scaling Latent Reasoning via Looped Language Models},
  author={Zhu, Q. and others},
  journal={arXiv preprint arXiv:2510.25741},
  year={2025}
}

\appendix

\section{Configuration}
\label{app:config}

\begin{table}[h]
\centering
\caption{Architecture and training configuration.}
\label{tab:config}
\begin{tabular}{lll}
\toprule
 & S0 (toy) & S1 (main) \\
\midrule
Data & TinyStories & FineWeb-Edu \\
Tokens & 0.4B & 12.9B \\
$d_{\text{model}}$ & 384 & 576 \\
Heads (KV heads) & 6 (2) & 9 (3) \\
FFN dim & 1024 & 1536 \\
Prelude / core / coda layers & 2 / 4 / 2 & 2 / 4 / 2 \\
Unique params (non-embedding) & 12.9M & 29.0M \\
Total params & 31.8M & 57.3M \\
$\bar{r}$ (mean loops) & 4 & 8 \\
$r_{\max}$ & 16 & 32 \\
Truncated-BPTT $k$ & 4 & 8 \\
Sequence length & 512 & 2048 \\
Peak LR & $1\times10^{-4}$ & $1\times10^{-4}$ \\
Optimizer & AdamW (0.9, 0.95), wd 0.1 & AdamW (0.9, 0.95), wd 0.1 \\
Schedule & cosine, 1\% warmup & cosine, 1\% warmup \\
Seeds & 2 of 3 & 1 \\
\bottomrule
\end{tabular}
\end{table}

Full configuration JSONs ship with the code release.

\section{Elasticity across training}
\label{app:elasticity}

\begin{table}[h]
\centering
\caption{S1 elasticity: validation loss vs inference loops (final checkpoint,
noise init, 64 windows).}
\label{tab:elasticity}
\begin{tabular}{cccc}
\toprule
loops $r$ & loss (nats) & ppl & tokens/s (batch 1) \\
\midrule
1 & 3.803 & 44.8 & 18336 \\
2 & 3.371 & 29.1 & 12952 \\
4 & 3.217 & 24.9 & 8186 \\
8 & \textbf{3.202} & 24.6 & 4727 \\
16 & 3.203 & 24.6 & 2577 \\
32 & 3.203 & 24.6 & 871 \\
\bottomrule
\end{tabular}
\end{table}

Provisional sweeps at 12, 28, 76, and 91 percent of training showed the same
monotone shape from the first checkpoint onward, with the whole curve descending
as training progressed; the $r=8$ to $r=16$ gap never opened at any checkpoint.
The S0 two-seed spread was at most 0.030 nats at any depth and 0.011 nats at
$r=16$.

\section{Convergence and allocation tables}
\label{app:tables}

\begin{table}[h]
\centering
\caption{Convergence diagnostic (S1 final, diagnostic run to 16 loops). Per loop: mean
$\mathrm{KL}(p_i \,\|\, p_{i-1})$, mean state change, and fraction of tokens
with KL $> 10^{-3}$.}
\label{tab:convergence}
\begin{tabular}{cccc}
\toprule
loop $i$ & mean KL & mean state change & percent tokens moving \\
\midrule
2 & $3.9\times10^{-1}$ & 24.7 & 99.8 \\
4 & $3.2\times10^{-2}$ & 7.85 & 96.3 \\
6 & $3.7\times10^{-3}$ & 2.60 & 72.7 \\
7 & $1.4\times10^{-3}$ & 1.52 & 33.0 \\
8 & $5.4\times10^{-4}$ & 0.90 & 10.0 \\
10 & $1.1\times10^{-4}$ & 0.34 & 1.4 \\
12 & $3.3\times10^{-5}$ & 0.14 & 0.4 \\
16 & $8.5\times10^{-6}$ & 0.030 & 0.1 \\
\bottomrule
\end{tabular}
\end{table}

\begin{table}[h]
\centering
\caption{Allocation: loss at matched average depth (S1 final, zero init, 128
windows). Uniform depth-8 quality bar $= 3.189$.}
\label{tab:allocation}
\begin{tabular}{cccc}
\toprule
avg depth & A0 uniform & A1 convergence exit & A2 trained router \\
\midrule
3 & 3.272 & \textbf{3.231} & 3.385 \\
4 & 3.203 & \textbf{3.201} & 3.313 \\
5 & 3.199 & \textbf{3.192} & 3.268 \\
6 & 3.196 & \textbf{3.190} & 3.236 \\
7 & 3.192 & \textbf{3.189} & 3.210 \\
\bottomrule
\end{tabular}
\end{table}

A1 reaches the depth-8 bar at 4.94 average loops (38 percent fewer); A2 needs
7.99.

\begin{table}[h]
\centering
\caption{Per-request cost curve (S1 final, uniform depth, batch 1).}
\label{tab:cost}
\begin{tabular}{cccc}
\toprule
depth $r$ & loss (nats) & tokens/s & relative throughput \\
\midrule
2 & 3.371 & 12952 & 2.7x \\
4 & 3.217 & 8186 & 1.7x \\
8 & 3.202 & 4727 & 1.0x \\
16 & 3.203 & 2577 & 0.55x \\
32 & 3.203 & 871 & 0.18x \\
\bottomrule
\end{tabular}
\end{table}

\section{Training stability notes}
\label{app:stability}

Two numerical issues encountered during S1 bring-up are documented here because
practitioners running mixed-precision training on this architecture are likely
to encounter them.

\paragraph{Mixed-dtype attention.}
Under bf16 autocast, the rotary embedding tables were built in fp32 (the
embedding output dtype), and multiplying the bf16 query and key projections by
fp32 tables promoted them to fp32 while the value projection remained bf16.
Fused scaled-dot-product attention kernels reject mixed dtypes, and a broad
exception handler in the training loop silently fell back to fp32 for the
remainder of the run, so runs labeled bf16 could execute in fp32 without any
visible error. The fix is one line, casting the rotary tables to the query dtype
before application, plus removing the silent fallback so that dtype errors
surface as errors. We recommend logging the observed dtypes of attention inputs
once at the first step of any mixed-precision run.

\paragraph{Non-finite step guard.}
The training loop skips the optimizer step when the loss or gradient norm is
non-finite, logs every skip with its step number, and aborts if skips exceed 0.1
percent of steps or three consecutive occurrences. On a healthy configuration
this guard never fires; treating any firing as a bug signal rather than noise to
suppress was, in our experience, the correct policy.

\end{document}